\documentclass[letterpaper, 10 pt, conference]{ieeeconf}  

\IEEEoverridecommandlockouts                              

\usepackage{amsmath,amsfonts}
\usepackage{array}
\usepackage[caption=false,font=normalsize,labelfont=sf,textfont=sf]{subfig}
\usepackage{textcomp}
\usepackage{stfloats}
\usepackage{url}
\usepackage{verbatim}
\usepackage{graphicx}
\usepackage{cite}
\usepackage{multirow}
\usepackage{lscape} 
\usepackage{tabularx} 
\usepackage{graphicx} 
\usepackage{comment}
\usepackage{color}
\usepackage{xcolor}
\usepackage{algorithm}
\usepackage[noend]{algpseudocode} 
\usepackage{amssymb,amsmath}
\usepackage{xspace}
\usepackage{hyperref}

\newtheorem{theorem}{Theorem}

\newtheorem{definition}{Definition}

\newtheorem{assumption}{Assumption}

\newcommand{\blue}{\color{black}}

\newcommand\abbrMAPF{MAPF\xspace}
\newcommand\abbrMCPF{MCPF\xspace}
\newcommand\abbrMCPFSum{MCPF-sum\xspace}
\newcommand\abbrMCPFMax{MCPF-max\xspace}
\newcommand\abbrHPP{HPP\xspace}
\newcommand\abbrMHPP{mHPP\xspace}
\newcommand\abbrMTSP{mTSP\xspace}
\newcommand\abbrTSP{TSP\xspace}
\newcommand\abbrTSPs{TSPs\xspace}

\newcommand\procName[1]{\textsl{#1}}

\newcommand\algname[1]{\textsf{#1}\xspace}
\newcommand\abbrDMS{\algname{DMS*}}
\newcommand\abbrMS{\algname{MS*}}
\newcommand\abbrAstar{\algname{A*}}
\newcommand\abbrMstar{\algname{M*}}

\newcommand\FrontierSet{\mathcal{F}}

\usepackage{ifthen}
\newboolean{shortver}
\setboolean{shortver}{false}

\begin{document}

\title{\LARGE \bf
DMS*: Minimizing Makespan for \\Multi-Agent Combinatorial Path Finding
}


\author{Zhongqiang Ren, Anushtup Nandy, Sivakumar Rathinam and Howie Choset
\thanks{Zhongqiang Ren is with Shanghai Jiao Tong University in China. (email: zhongqiang.ren@sjtu.edu.cn). Anushtup Nandy and Howie Choset are with Carnegie Mellon University in USA. (email: anandy@andrew.cmu.edu; choset@andrew.cmu.edu). Sivakumar Rathinam is with Texas A\&M University in USA (email: srathinam@tamu.edu).}
}



\maketitle

		\thispagestyle{plain}
		\pagestyle{plain}
		\pagenumbering{arabic}

	\begin{abstract}
		Multi-Agent Combinatorial Path Finding (\abbrMCPF) seeks collision-free paths for multiple agents from their start to goal locations, while visiting a set of intermediate target locations in the middle of the paths.
\abbrMCPF is challenging as it involves both planning collision-free paths for multiple agents and target sequencing, i.e., solving traveling salesman problems to assign targets to and find the visiting order for the agents.
Recent work develops methods to address \abbrMCPF while minimizing the sum of individual arrival times at goals.
Such a problem formulation may result in paths with different arrival times and lead to a long makespan, the maximum arrival time, among the agents.
This paper proposes a min-max variant of \abbrMCPF, denoted as \abbrMCPFMax, that minimizes the makespan of the agents.
While the existing methods (such as \abbrMS) for \abbrMCPF can be adapted to solve \abbrMCPFMax, we further develop two new techniques based on \abbrMS to defer the expensive target sequencing during planning to expedite the overall computation.
We analyze the properties of the resulting algorithm Deferred \abbrMS (\abbrDMS), and test \abbrDMS with up to 20 agents and 80 targets.
We demonstrate the use of \abbrDMS on differential-drive robots.

	\end{abstract}

	
	\graphicspath{{./figure/}}
	
	\section{Introduction}\label{dms:sec:intro}
	
Multi-Agent Path Finding (\abbrMAPF) seeks a set of collision-free paths for multiple agents from their respective start to goal locations.
This paper considers a generalization of \abbrMAPF called Multi-Agent Combinatorial Path Finding (\abbrMCPF), where the agents need to visit a pre-specified set of intermediate target locations before reaching their goals.
\abbrMAPF and \abbrMCPF arise in applications such as logistics~\cite{wurman2008coordinating}.
For instance, factories use a fleet of mobile robots to visit a set of target locations to load machines for manufacturing.
These robots share a cluttered environment and follow collision-free paths.
In such settings, \abbrMAPF problems and their generalizations naturally arise to optimize operations.

\begin{figure}[tb]
\centering
\includegraphics[width=0.88\linewidth]{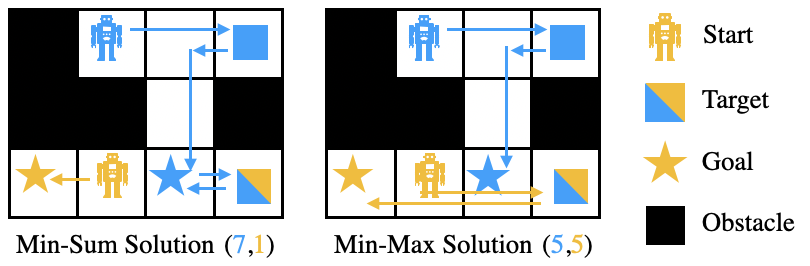}
\vspace{-3mm}
\caption{
    \abbrMCPFMax and \abbrMCPFSum. \abbrMCPFMax seeks a set of collision-free paths while minimizing the maximum arrival time of the agents.
    The color of a target or goal indicates the assignment constraints, i.e., the subset of agents that are eligible to visit that target or goal.
    }
\vspace{-5mm}
\label{dms:fig:fig1}
\end{figure}

\abbrMCPF is challenging as it involves both collision avoidance among the agents as in \abbrMAPF, and target sequencing, i.e., solving Traveling Salesman Problems (TSPs)~\cite{Applegate:2007,bektas2006multiple} to specify the assignment and visiting orders of targets for all agents.
Both the TSP and the \abbrMAPF are NP-hard to solve to optimality~\cite{yu2013structure_nphard,Applegate:2007}, and so is \abbrMCPF.
A few methods were developed~\cite{ren21ms,ren23cbssTRO} to address \abbrMCPF, and they often formulate the problem as a min-sum optimization problem, denoted as \abbrMCPFSum, where the objective is to minimize the sum of individual arrival times.
Such a formulation may result in an ensemble of paths where some agents arrive early while others arrive late, which leads to long execution times before all agents finish their paths.
This paper thus proposes a min-max variant of \abbrMCPF (Fig.~\ref{dms:fig:fig1}), denoted as \abbrMCPFMax, where the objective is to minimize the maximum arrival time, which is also called the makespan, of all agents.

To solve \abbrMCPFMax, we first adapt our prior \abbrMS algorithm~\cite{ren21ms}, which was designed for \abbrMCPFSum, to address \abbrMCPFMax.
Then, we further develop two new techniques to expedite the planning, and we call the resulting new algorithm Deferred \abbrMS (\abbrDMS).
Specifically, the existing \abbrMS is a heuristic search approach (such as \abbrAstar) by iteratively generating, selecting and expanding states to construct partial solution paths from the initial state to the goal state.
\abbrMS uses Traveling Salesman Problem (\abbrTSP) algorithms to compute target sequences for the agents.
When solving the \abbrTSP, agent-agent collision are ignored, and the cost of resulting target sequences are thus lower bounds of the true costs to reach the goals.
Therefore, the cost of target sequences provides an admissible heuristic to guide the state selection and expansion as in \abbrAstar.
Furthermore, \abbrMS leverages the idea in \abbrMstar~\cite{wagner2015subdimensional} to first use the target sequences to build a low-dimensional search space, and then grow this search space by coupling agents together for planning only when collision happens.
By doing so, \abbrMS interleaves \abbrTSP (target sequencing) and \abbrMAPF (collision resolution) techniques using a heuristic search approach, and provides completeness and solution optimality.

The first technique developed in this paper is applicable to \abbrMS for both \abbrMCPFMax and \abbrMCPFSum.
When expanding a state, a set of successor states are generated, and for each of them, \abbrMS needs to invoke the \abbrTSP solver to find the target sequence and the heuristic value of this successor state.
Since the number of successor states can be large for each expansion, \abbrMS needs to frequently invokes the \abbrTSP solver which slows down the computation.
To remedy this issue, for each generated successor, we first use a fast-to-compute yet roughly estimated cost-to-go as the heuristic value, and defer calling \abbrTSP solver for target sequencing until that successor is selected for expansion.

The second technique is only applicable to \abbrMS when solving \abbrMCPFMax, and does not work for \abbrMCPFSum.
Since the goal here is to minimize the makespan, during the search, agents with non-maximum arrival time naturally have ``margins'' in a sense that they can take a longer path without worsening the makespan of all agents.
We take advantage of these margins to let agents re-use their previously computed target sequences and defer the expensive calls of \abbrTSP solvers until the margin depletes.

To verify the methods, we conduct both simulation in various maps with up to 20 agents and 80 targets, as well as a simple real robot experiment.
The simulation shows that: (i) \abbrDMS finds paths that are up to 50\% cheaper than an iterative greedy baseline method, and (ii) the new techniques in \abbrDMS help triple the success rates and reduce the average runtime to solution comparing to \abbrMS.
The robot experiments show that the planned path are executable, and inspire future work.

	\section{Related Work}\label{dms:sec:related}
        
\noindent\textbf{Multi-Agent Path Finding} algorithms fall on a spectrum from coupled~\cite{standley2010finding} to decoupled~\cite{silver2005hca}, trading off completeness and optimality for scalability.
In the middle of this spectrum lie the dynamically-coupled methods such as \abbrMstar~\cite{wagner2015subdimensional} and \algname{CBS} \cite{sharon2015conflict}, which begin by planning for each agent a shortest path from the start to the goal ignoring any potential collision with the other agents, and then couple agents for planning only when necessary to avoid agent-agent collision.

\noindent\textbf{Traveling Salesman Problems}  determine both the assignment and visiting order of the targets for the agents, where there are multiple intermediate targets to visit.
For a single agent, the Traveling Salesman Problem (\abbrTSP) seeks a shortest tour that visits every vertex in a graph, and is a well-known NP-hard problem~\cite{Applegate:2007}.
Closely related to \abbrTSP, the Hamiltonian Path Problem (\abbrHPP) requires finding a shortest path that visits each vertex in the graph from a start vertex to a goal vertex.
The multi-agent version of the \abbrTSP and \abbrHPP (denoted as \abbrMTSP and \abbrMHPP, respectively) are more challenging since the vertices in the graph must be allocated to each agent in addition to finding the optimal visiting order of vertices.
We refer to all these problems simply as \abbrTSPs.
Different methods have been developed~\cite{Applegate:2007,helsgaun2009general,christofides1976worst} to solve \abbrTSPs, trading off solution optimality for runtime efficiency.
This paper does not develop new \abbrTSPs solvers and leverage the existing ones.

\noindent\textbf{Target Assignment, Sequencing and Path Finding} were recently combined in different ways~\cite{honig2018conflict,ma2016optimal,nguyen2019generalized,surynek2021multi,ren23cbssTRO,ren21ms,zhong2022optimal,zhang2022multi}.
Most of them either consider target assignment only (without the need for computing visiting orders of targets)~\cite{honig2018conflict,ma2016optimal,nguyen2019generalized}, or consider the visiting order only given that each agent is pre-allocated a set of targets~\cite{surynek2021multi,zhong2022optimal,zhang2022multi}.
Our prior work~\cite{ren21ms,ren23cbssTRO} seeks to handle the challenge in target assignment and ordering and the collision avoidance in \abbrMAPF simultaneously.
These work uses the \abbrMCPFSum formulation and the developed planners minimize the sum of individual arrival times, while this paper investigates \abbrMCPFMax.
We do not extend our prior method CBSS~\cite{ren23cbssTRO} for \abbrMCPFSum to \abbrMCPFMax, because CBSS requires solving a min-sum K-best sequencing problem, and it is not obvious how to handle the min-max variant of this K-best sequencing problem.

	\section{Problem}\label{dms:sec:problem}
	Let the index set $I = \{1,2,\dots,N\}$ denote a set of $N$ agents.
All agents share a workspace that is represented as an undirected graph $G^W=(V^W,E^W,c^W)$, where $W$ stands for workspace.
Each vertex $v\in V^W$ represents a possible location of an agent.
Each edge $e=(u,v) \in E^W \subseteq V^W\times V^W$ represents an action that moves an agent between $u$ and $v$.
$c^W: E^W\rightarrow (0,\infty)$ maps an edge to its positive cost value.
In this paper, the cost of an edge is equal to its traversal time, and each edge has a unit cost.\footnote{\blue
Here, the edge cost is the same as the edge traversal time, which is one unit for each edge. When the traversal time of edges are not unitary, it leads to continuous-time MAPF, and we refer the reader to~\cite{andreychuk2022multi,ren21loosely}.}

Let the superscript $i \in I$ over a variable denote the specific agent to which the variable belongs (e.g. $v^i\in V^{W}$ means a vertex corresponding to agent $i$).
Let $v_o^i,v_d^i\in V^{W}$ denote the \emph{initial} (or original) vertex and the \emph{goal} (or destination) vertex of agent $i$ respectively.
Let $V_o,V_d \subset V^W$ denote the set of all initial and goal vertices of the agents respectively, and
let $V_t \subset V^W\backslash\{V_o\cup V_d\}$ denote the set of \emph{target} vertices.
For each target $v\in V_t$, let $f_A(v)\subseteq I$ denote the subset of agents that are eligible to visit $v$; these sets are used to formulate the (agent-target) \emph{assignment constraints}.\footnote{
An agent $i$ ``visits'' a target $v\in V_t$ means (i) there exists a time $t$ such that agent $i$ occupies $v$ along its path, and (ii) the agent $i$ claims that $v$ is visited. If a target $v$ is in the middle of the path of $i$ and $i$ does not claim $v$ is visited, then $v$ is not considered as visited. A visited target $v$ can appear in the path of another agent. When we say an agent or a path ``visits'' a target, we always mean the agent ``visits and claims'' the target. {\blue The assignment constraints do not forbid any agent $j \notin f_A(v)$ to use $v$ in its path, and only forbid agent $j$ to claim visiting $v$.}
}

Let $\pi^i(v^i_{1}, v^i_{\ell})$ denote a path for agent $i$ between vertices $v^i_{1}$ and $v^i_{\ell}$, which is a list of vertices $(v^i_{1},v^i_{{2}},\dots,v^i_{\ell})$ in $G^W$ with $(v^i_{k}, v^i_{k+1})\in E^W, k=1,2,\cdots,\ell-1$.
Let $g(\pi^i(v^i_{1}, v^i_\ell))$ denote the cost of the path, which is the sum of the costs of all edges present in the path: $g(\pi^i(v^i_{1}, v^i_{\ell})) = \Sigma_{j=1,2,\dots,{\ell-1}} c^W(v^i_{{j}}, v^i_{{j+1}})$.

All agents share a global clock and start to move along their paths from time $t=0$.
Each action of the agents, either wait or move along an edge, requires one unit of time.
Any two agents $i,j \in I$ are in \emph{conflict} if one of the following two cases happens.
The first case is a \emph{vertex conflict} $(i,j,v,t)$ where two agents $i,j\in I$ occupy the same vertex $v$ at the same time $t$.
The second case is an \emph{edge conflict} $(i,j,e,t)$, where two agents $i,j\in I$ go through the same edge $e$ from opposite directions between times $t$ and $t+1$.

\begin{definition}[\abbrMCPFMax Problem]\label{dms:def:problem_mcpfMax}
The MCPF with Min-Max Objective (\abbrMCPFMax) seeks to find a set of conflict-free paths for the agents such that (1) each target $v \in V_t$ is visited at least once by some agent in $f_A(v)$,
(2) the path for each agent $i \in I$ starts at its initial vertex and terminates at a unique goal vertex $v^i_d \in V_d$ such that $i \in f_A(v^i_d)$,
and (3) the maximum of the cost of all agents' paths (i.e., $\max_{i\in I}g(\pi^i(v^i_o,v^i_d)))$ reaches the minimum.
\end{definition}

	\section{Method}\label{dms:sec:method}
	
\begin{figure*}[tb]
\centering
\includegraphics[width=\linewidth]{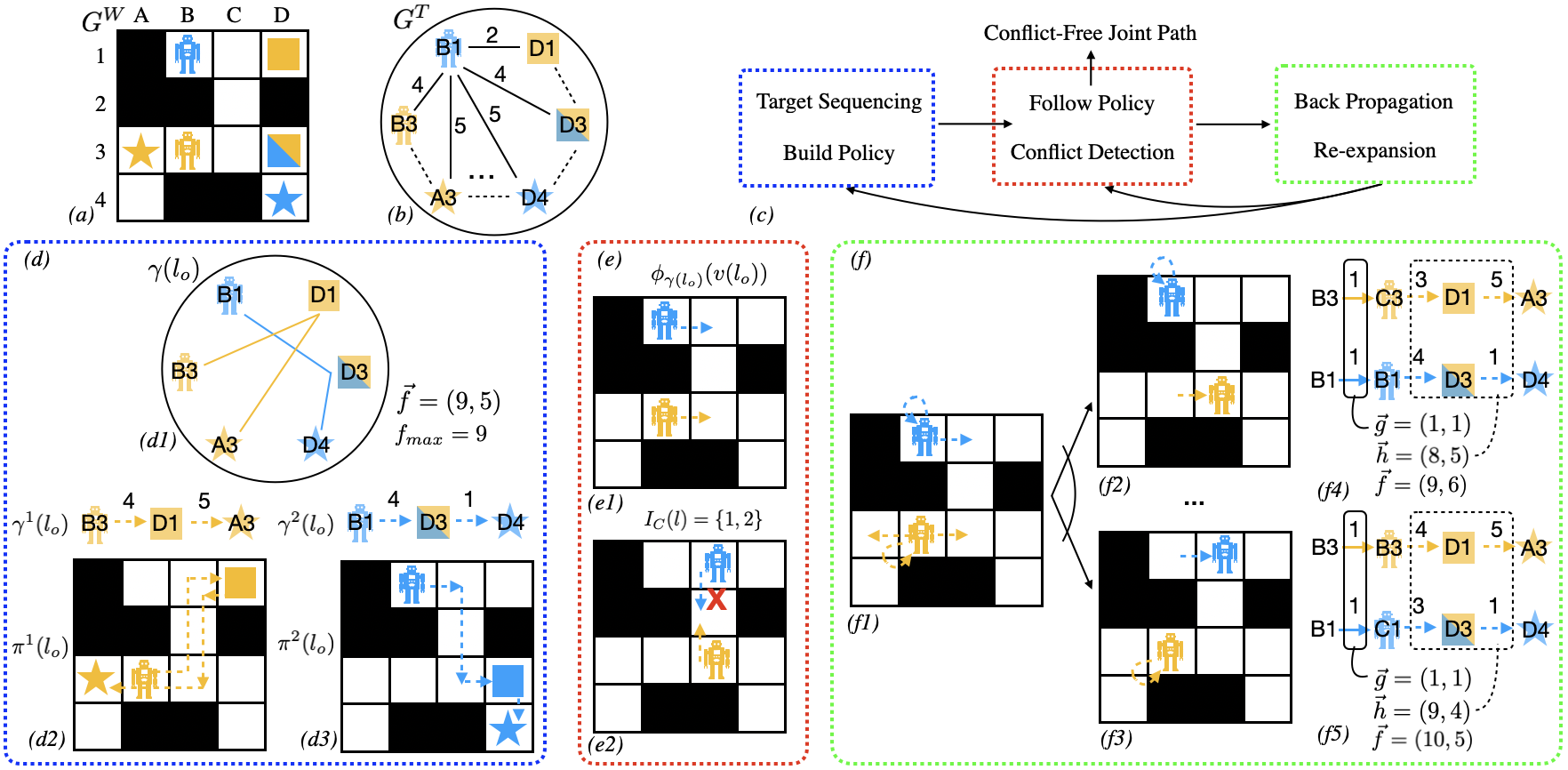}
\vspace{-5mm}
\caption{
    An illustration of \abbrDMS and related concepts.
    (a) shows the workspace graph $G^W$.
    (b) shows the target graph $G^T$ where each edge in $G^T$ corresponds to a minimum cost path in $G^W$ between the respective vertices.
    (c) shows the workflow of \abbrDMS in Alg.~\ref{dms:alg:dms}.
    (d) shows \abbrDMS first ignores any agent-agent conflict and solves a corresponding \abbrMHPP, which provides a joint sequence $\gamma(l_o)$ (d1).
    This joint sequence $\gamma(l_o)$ can be converted to a joint path $\pi(l_o)$ (d2,d3), whose corresponding makespan $f_{max}$ is 9.
    (e) The joint path $\pi(l_o)$ leads to a policy $\phi_{\gamma(l_o)}$ that maps a joint vertex to the next joint vertex along $\pi(l_o)$. Since conflicts are ignored in this policy, agents may run into conflict (e2). When a conflict is detected, the subset of agents that are in conflict $I_C(l)$ is back propagated to the ancestor labels.
    (f) After the back propagation, these ancestor labels and re-opened and re-expanded while considering all possible actions of all agents in conflict (f1,f2,f3).
    During the re-expansion, for each of the successors, \abbrDMS first uses a fast-to-compute yet roughly estimated cost-to-go as the heuristic of the generated labels to avoid solving a \abbrMHPP. For both labels (f2) and (f3), this rough heuristic is $(9,5)-(1,1)=(8,4)$, where $(9,5)$ is the heuristic of (f1), the parent of (f2) and (f3), and $(1,1)$ means each agent can move at most one step towards their goals.
    When either (f2) or (f3) is popped from OPEN and before being expanded, \abbrDMS re-computes a new heuristic and checks if the popped label should be expanded or re-added to OPEN for future expansion.
    To obtain this new heuristic, \abbrDMS first attempts to let the agents follow the previously computed target sequences of the parent label and check if this worsens the makespan.
    For the successor shown in (f2), the resulting makespan is 9 (f4), which is no worse than the previous makespan 9, and \abbrDMS will not call \abbrMHPP solver to save computational effort.
    For the successor shown in (f3), the resulting makespan is 10 (f5), which is worse than the previous makespan 9, and \abbrDMS have to call \abbrMHPP solver to find a new joint sequence from that successor (f3) to the goals.
    Finally, with the joint sequence, new policy for the successors can be built and the search continues as in (e).
    }
\vspace{-5mm}
\label{dms:fig:eg}
\end{figure*}

This section begins with an example in Fig.~\ref{dms:fig:eg} to illustrate the planning process of \abbrDMS.
{\blue
Then, Sec.~\ref{dms:sec:method:concepts} introduces some concepts that will be used during the search process, before the presentation of the search algorithm in Sec.~\ref{dms:sec:method:alg}.
}

\subsection{Concepts and Notations}\label{dms:sec:method:concepts}
\subsubsection{Joint Graph}
Let $\mathcal{G^W}=(\mathcal{V^W},\mathcal{E^W}) = \underbrace{G^W \times G^W \times \dots \times G^W}_{\text{$N$ times}}$\vspace{0.5em} denote the \emph{joint graph} of the agents which is the Cartesian product of $N$ copies of $G^W$, where each $v \in \mathcal{V^W}$ represents a \emph{joint vertex} and $e \in \mathcal{E^W}$ represents a \emph{joint edge} that connects a pair of joint vertices.
Let $v_o = (v^1_o,v^2_o,\cdots,v^N_o)$ denote the initial joint vertex, which contains the initial vertices of all the agents.
Let $\pi(u,v), u,v \in \mathcal{V^W}$ denote a \emph{joint path}, which is a tuple of $N$ (individual) paths of the same length, i.e., $\pi(u,v) = (\pi^1(u^1,v^1), \cdots, \pi^N(u^N,v^N))$.
{\blue
A joint path $\pi$ can be viewed as a list of joint vertices $(v_0,v_1,\cdots,v_\ell)$, and the time for the agents to arrive at each joint vertex $v \in \pi$ is specified by the index of $v$ in $\pi$.
Given two subsequent joint vertices $u,v$ along a joint path $\pi$, let $\procName{CheckConflict}(u,v)$ denote a procedure that checks for vertex conflicts at $u,v$ and edge conflicts during the transition from $u$ to $v$ among all pairs of agents, and \procName{CheckConflict} returns a set $I_C \subseteq I$ of agents that are in conflict.} 
We use a ``path'' to denote an ``individual path''.
\abbrDMS searches the joint graph $\mathcal{G^W}$ for a joint path that solves the \abbrMCPFMax.

\subsubsection{Binary Vector}
For any $v\in \mathcal{V^W}$, there can be multiple joint paths, e.g. $\pi_1(v_o,v),\pi_2(v_o,v)$, from $v_o$ to $v$ with different sets of targets visited, and we need to differentiate between them.
First, without losing generality, let all targets in $V_t$ be arranged as an ordered list $V_t=\{u_m, m = 1,2,\cdots,M\}$, where subscript $m$ indicates the index of a target in this list.\footnote{In this paper, for a vector related to targets (e.g. a binary vector $\vec{a}$ of length $M$), we use subscripts (e.g. $a_m$) to indicate the elements in the vector. For a vector or joint vertex that is related to agents (e.g. $\vec{g}$ of length $N$, or $v \in \mathcal{V^W}$), we use a superscript (e.g. $i$ in $g^i,v^i$) to indicate the element in the vector or joint vertex corresponding to an agent.}
Let $\Vec{a} \in \{0,1\}^{M}$ denote a \emph{binary vector} of length $M$ that indicates the visiting status of all targets in $V_t$ {\blue by a joint path during the search}, where the $m$-th component of $\Vec{a}$ is denoted as $a_m$, and $a_m=1$ if $u_m$ is visited, and $a_m=0$ otherwise.

\subsubsection{Label}
Let $l=(v,\Vec{a},\Vec{g})$ denote a \emph{label}, where $v$ is a joint vertex, $\Vec{a}$ is a binary vector of length $M$, and $\Vec{g}$ is a \emph{cost vector} of length $N$.
Here, each component $g^i, i\in I$ is the path cost of agent $i$.
{\blue During the search}, each label $l$ identifies a joint path from $v_o$ to $v$ that visits a subset of targets as described by $\Vec{a}$ and with path cost $\Vec{g}$.
Given $l$, let $v(l),\Vec{a}(l), \Vec{g}(l)$ denote the corresponding component in $l$, and let $v^i(l)$ denote the vertex of agent $i$ in $v(l)$, $i\in I$.
Let $g_{max}(l):=\max_{i\in I}\Vec{g}(l)$ denote the maximum path cost over all agents in $\Vec{g}(l)$.
To solve the \abbrMCPFMax in Def.~\ref{dms:def:problem_mcpfMax}, the planner searches for a label $l$, whose corresponding joint path leads all agents to visit all targets and eventually reaches the goals, and $g_{max}(l)$ is the \emph{objective value} to be minimized.

\subsubsection{Label Comparison}
To compare two labels at the same joint vertex, we compare both $\vec{a}$ and $\vec{g}$.

\begin{definition}[Binary Dominance]
For any two binary vectors $\Vec{a}$ and $\Vec{b}$, $\Vec{a}$ dominates $\Vec{b}$ ($\Vec{a} \succeq_b \Vec{b}$), if both the following conditions hold: (i) $\forall m \in \{1,2,\cdots,M\}$, ${a}_m\geq {b}_m$; (ii) $\exists m \in \{1,2,\cdots,M\}$, ${a}_m> {b}_m$.
\end{definition}
Intuitively, $\Vec{a}\succeq_b\Vec{b}$ if $\Vec{a}$ visits all targets that are visited in $\Vec{b}$, and $\Vec{a}$ visits at least one more target than $\Vec{b}$.
Two binary vectors are equal to each other ($\Vec{a}=\Vec{b}$) if both vectors are component-wise same to each other.
\begin{definition}[Label Dominance]\label{dms:def:labelDom}
For two labels ${l}_1$, ${l}_2$ with $v(l_1)=v(l_2)$, $l_1$ dominates $l_2$ ($l_1 \succeq_l l_2$) if either (i) $\Vec{a}(l_1) \succeq_b \Vec{a}(l_2)$, $g_{max}(l_1) \leq g_{max}(l_2)$; or (ii) $\Vec{a}(l_1) = \Vec{a}(l_2)$, $g_{max}(l_1) < g_{max}(l_2)$ holds.
\end{definition}
Intuitively, if $l_1 \succeq_l l_2$, then the joint path identified by $l_1$ is guaranteed to be better than the joint path identified by $l_2$.
If $l_1$ does not dominate $l_2$, {\blue $l_2$ is then non-dominated by $l_1$.}
Any two labels are non-dominated (with respect to each other) if each of them is non-dominated by the other.
Two labels are said to be \emph{equal to} (or \emph{same to}) each other (notationally $l_1=l_2$) if $v(l_1)=v(l_2),a(l_1)=a(l_2),g_{max}(l_1)=g_{max}(l_2)$.
There is no need to compare $\Vec{g}(l_1)$ and $\Vec{g}(l_2)$ when comparing labels $l_1,l_2$ since the problem in Sec.~\ref{dms:sec:problem} seeks to minimize the maximum path cost over the agents.
Finally, for each joint vertex $v\in \mathcal{V^W}$, let $\FrontierSet(v)$ denote a set of labels that are non-dominated to each other during the search.

\subsubsection{Target Sequencing}

{\blue
Let $\gamma^i=\{v^i_o,u_1,u_2,\cdots,u_k,v^i_d\}$ denote an (individual) target sequence, where each $u_j, j=1,2,\cdots,k$ is a target vertex (i.e., $u_j \in V_t$).
}
Let $\gamma=\{\gamma^1, \gamma^2, \cdots, \gamma^N \}$ denote a \emph{joint sequence}, which specify the assignment and visiting order of all targets for all agents.
Given $l$, $\vec{a}(l)$ specifies the set of targets that are visited and unvisited.
We introduce the notation $\gamma(l)=\{\gamma^1(l), \gamma^2(l), \cdots, \gamma^N(l)\}$, a joint sequence \emph{based on} $l$ in the sense that $\gamma(l)$ visits all unvisited targets in $\Vec{a}(l)$: (i) each $\gamma^i(l) \in \gamma(l)$ starts with $v^i(l)$ ({\blue $v^i(l)$ is not necessarily an origin vertex}), visits a set of unvisited targets $\{u_m\}$ (where the $m$-th component of $\vec{a}(l)$ is zero), and ends with a goal vertex $v^i_d$; (ii) all $\gamma^i(l), i\in I$ together visits all unvisited targets in $\vec{a}(l)$.
{\blue Intuitively, $\gamma(l)$ is a joint sequence that is meant to complete the joint path represented by $l$.}

\subsubsection{Target Graph}
Let $\pi^*(u,v),u,v\in G^W$ denote a minimum-cost path between $u,v$ in $G^W$, and let $c_{\pi^*}(u,v)$ denote the cost of path $\pi^*(u,v)$.
The cost of a target sequence $c(\gamma^i(l))$ is equal to the sum of $c_{\pi^*}(u,v)$ for any two adjacent vertices $u,v$ in $\gamma^i(l)$.
Given $l$, to find $\gamma(l)$, a corresponding min-max multi-agent Hamiltonian path problem (\abbrMHPP) needs to be solved as follows.
First, a target graph $G^T=(V^T,E^T,c^T)$ is created based on $G^W$.
The $V^T$ includes the current vertices of agents $v(l)$, the unvisited targets $\{u_m \in V_t | a_m(l) = 0\}$ and goals $V_d$.
$G^T$ is fully connected and $E^T = V^T \times V^T$.
The edge cost in $G^T$ of any pair of $u,v\in V^T$ is denoted as $c^T(u,v)$, which is the cost of a minimum cost path $\pi^*(u,v)$ in the $G^W$.
An example $G^T$ is shown in Fig.~\ref{dms:fig:eg}(b).
A target sequence $\gamma^i(l)$ for an agent $i$ is a path in $G^T$, and a $\gamma(l)$ is a set of paths that starts from $v(l)$, visits all unvisited targets in $V_t$ as in $\vec{a}(l)$, and ends at goals $V_d$, while satisfying the assignment constraints.
The procedure \procName{SolveMHPP}$(l)$ can be implemented by various existing algorithms for \abbrMHPP.

\subsubsection{Heuristic and Policy}
Given $\gamma(l)$, let $h^i(l):=c(\gamma^i(l))$ be a \emph{heuristic} value.
The vector $\vec{h}:=\{h^i(l) | i\in I\}$ provides an estimate of the cost-to-go for each agent $i \in I$.
When \procName{SolveMHPP}($l$) solves the \abbrMHPP to optimality, since conflicts are ignored along the target sequences, the corresponding $\Vec{h}$ provides lower bounds of the cost-to-go for all agents, which is an admissible heuristic for the search.
For any label $l$, let $\Vec{f}(l) := \vec{g}(l) + \vec{h}(l)$ be the $f$-vector of $l$, which is an estimated cost vector of the entire joint path from $v_o$ to goals for all agents by further extending the joint path represented by $l$.
Let $f_{max}(l):= \max_{i \in I}(f^i(l))$ denote the maximum component in $\Vec{f}(l)$, which provides an estimate of the objective value related to $l$. 

Given $l$ and $\gamma(l)$, a \emph{joint policy} $\phi_{\gamma(l)}$ is built out of $\gamma(l)$, mapping one label to another as follows.
First, a joint path $\pi$ is built based on $\gamma(l)$ by replacing any two subsequent vertices $u,v \in \gamma^i(l), i\in I$ with a corresponding minimum cost path $\pi^*(u,v)$ in $G^W$.
Then, along this joint path $\pi=(v_0, v_1, v_2, \cdots, v_\ell)$, all agents move from $v_k$ to $v_{k+1}$, $k=0,1,\cdots,\ell-1$.
The corresponding binary vectors $\vec{a}_k$ for each $v_k \in \pi$ are built by first making $\vec{a}_0 = \vec{a}(l)$, and then updating $\vec{a}_k, k=1,2,\cdots,\ell$ based on $\vec{a}_{k-1}$ and $v_k$ by checking if $v_k$ visits any new targets.
The corresponding cost vectors $\vec{g}_k$ are computed similarly as $a_k$ by starting from $\vec{g}_0=\vec{g}(l)$.
As a result, a joint policy $\phi_{\gamma(l)}$ is built by mapping one label $l=(v,\Vec{a},\Vec{g})$ to the next label $l'=(v',\Vec{a}',\Vec{g}')$ along the target sequence $\gamma(l)$.
For a vertex $v^i$, let $\phi^i_{\gamma(l)}(v^i)$ denote the next vertex of agent $i$ in the joint policy $\phi_{\gamma(l)}$.
An example of $\phi^i_{\gamma(l)}$ is shown in Fig.~\ref{dms:fig:eg}(d).
A label $l$ is \emph{on-policy} if its next label is known in $\phi_{\gamma(l')}$.
Otherwise, $l$ is \emph{off-policy}, i.e., the next label is unknown and a mHPP needs to be solved for $l$ to find the policy $\phi_{\gamma(l)}$. 
Let $\gamma(l),\vec{h}(l),\phi_{\gamma(l)} \gets $\procName{SolveMHPP}($l$) denote the process of computing the joint sequence, heuristic values and joint policy.

\subsection{\abbrDMS Algorithm}\label{dms:sec:method:alg}

\setcounter{topnumber}{8}
\setcounter{bottomnumber}{8}
\setcounter{totalnumber}{8}

\begin{algorithm}[t]
\small
    \caption{Pseudocode for \abbrDMS}\label{dms:alg:dms}
    \begin{algorithmic}[1]
    \State{$l_o \gets (v_o, \Vec{a}=0^M, \Vec{g}=0^N)$}\label{dms:alg:dms:lineInitBegin}
    \State{$parent(l_o) \gets NULL$, $f_{temp}(l_o) \gets 0$}
    
    \State{add $l_o$ to OPEN with $f_{temp}(l_o)$ as the priority}\label{dms:alg:dms:lineInitEnd}
    \State{add $l_o$ to $\FrontierSet(v_o)$}
    
    \While{OPEN is not empty} \label{dms:alg:dms:lineWhileBegin}
    
    
    \State{$l=(v,\Vec{a},\Vec{g}) \gets$ OPEN.pop() }

    \State{\procName{TargetSeq}($parent(l)$, $l$, $I_C(parent(l))$)}\label{dms:alg:dms:lineTargetSeq}
    \State{$f_{max}(l) \gets \max_{i\in I} \{ \vec{g}(l) +  w \cdot \vec{h}(l)$\} }\label{dms:alg:dms:lineFmax}
    \If{$f_{max}(l) > f_{temp}(l)$}\label{dms:alg:dms:lineFmaxCheck}
    \State{$f_{temp}(l) \gets f_{max}(l)$}
    \State{add $l$ to OPEN with $f_{temp}(l)$ as the priority}
    \State{\textbf{continue}}
    \EndIf

    \State{\textbf{if} $\procName{CheckSuccess}(l)$  \textbf{then}}
    \State{\indent \textbf{return} \procName{Reconstruct}($l$)}
    
    \State{$L_{succ} \gets$ \textsl{GetSuccessors}($l$) } 
    \ForAll{$l' \in L_{succ}$}
    
    \State{$I_C(l') \gets \procName{CheckConflict}(v(l),v(l'))$}\label{dms:alg:dms:lineDetectConflict}
    \State{\procName{BackProp}($l$, $I_C(l')$)}
    \If{$I_C \neq \emptyset$ \textbf{continue}}
    \EndIf
    
    \If{\procName{IsDominated}($l'$)}
    \State{\procName{DomBackProp($l,l'$)}}
    \State{\textbf{continue}}
    \EndIf

    
    \State{$f_{temp}(l') \gets \max_{i\in I} \{ \Vec{g}(l') + w \cdot \procName{SimpleHeu}(l') \}$}\label{dms:alg:dms:lineGetF}
    \State{$parent(l') \gets l$}
    \State{add $l'$ to $\FrontierSet(v(l'))$ and back\_set($l'$)}
    \State{add $l'$ OPEN with $f_{temp}(l')$ as the priority}
    \EndFor
    \EndWhile \label{dms:alg:dms:lineWhileEnd}
    \State{\textbf{return} Failure (no solution)}
    \end{algorithmic}
\end{algorithm}

\begin{algorithm}[t]
\small
\caption{Pseudocode for \procName{TargetSeq}($l$, $l'$, $I_C(l)$)}\label{dms:alg:targetSeq}
\begin{algorithmic}[1]
    \If{$l'$ is on-policy}
    \State{$\gamma(l') \gets \gamma(l)$, \textbf{return}}
    \EndIf

    \State{$\gamma(l') \gets \gamma(l)$}\label{dms:alg:targetSeq:lineCopySeq}
    \State{Compute $\phi_{\gamma(l')}$ and $\Vec{h}(l')$ based on $\gamma(l')$}\label{dms:alg:targetSeq:lineCopySeqThen}
    
    \If{$\exists i \in I_C(l), g^i(l') + h^i(l') > f_{max}(l')$}\label{dms:alg:targetSeq:lineCheckFmax}
    \State{$\gamma(l'),\Vec{h}(l'),\phi_{\gamma(l')} \gets \procName{SolveMHPP}(l')$}\label{dms:alg:targetSeq:lineCheckFmaxEnter}
    \EndIf
    \State{\textbf{return}}
\end{algorithmic}
\end{algorithm}

\begin{algorithm}[t]
\small
\caption{Pseudocode for BackProp($l,I_C(l')$)}\label{dms:alg:backprop}
\begin{algorithmic}[1]
    
    
    \If{$I_C(l') \nsubseteq I_C(l)$}
    \State{$I_C(l) \gets I_C(l') \bigcup I_C(l)$}
    \If{$l \notin$ OPEN}{ add $l$ to OPEN}
    \EndIf
    \ForAll{$l'' \in$ back\_set($l$)}
    \State{\procName{BackProp}($l''$, $I_C(l)$)}\label{dms:alg:backprop:lineRecursive}\label{dms:alg:backprop:lineKBPDecrease}
    \EndFor
    \EndIf
\end{algorithmic}
\end{algorithm}

\begin{algorithm}[t]
\small
\caption{Pseudocode for DomBackProp($l,l'$)}\label{dms:alg:dom_back_prop}
\begin{algorithmic}[1]
    \ForAll{$l'' \in \FrontierSet(v(l'))$}
    \If{$l'' \succeq_l l' \text{ or } l'' = l'$}
    \State{\procName{BackProp}($l$, $I_C(l'')$)}
    \State{add $l$ to back\_set($l''$)}
    \EndIf
    \EndFor
\end{algorithmic}
\end{algorithm}

To initialize (Lines~\ref{dms:alg:dms:lineInitBegin}-\ref{dms:alg:dms:lineInitEnd}), \abbrDMS creates an initial label $l_o$ and calls \procName{SolveMHPP} to compute $\gamma(l_o),\vec{h}(l_o),\phi_{\gamma(l_o)}$ for $l_o$.
For each label $l$, \abbrDMS uses two $f$-values: $f_{temp}(l)$ and $f_{max}(l)$, where $f_{temp}(l)$ is a fast-to-compute yet roughly estimated cost-to-go, which does not require computing any joint sequence from $l$ to the goals; and $f_{max}(l)$ is an estimated cost-to-go based on a joint sequence, which is computationally more expensive to obtain than $f_{temp}(l)$.
{\blue Similarly to \abbrAstar~\cite{astar}, let OPEN denote a priority queue storing labels and prioritizing labels based on their $f$-values from the minimum to the maximum.}
In Alg.~\ref{dms:alg:dms}, we point out which $f$-value (either $f_{temp}$ or $f_{max}$) is used when a label is added to OPEN.
Finally, $l_o$ is added to $\FrontierSet(v_o)$ since $l_o$ is non-dominated by any other labels at $v_o$, and $l_o$ is added to OPEN for future search.

In each iteration (Lines~\ref{dms:alg:dms:lineWhileBegin}-\ref{dms:alg:dms:lineWhileEnd}), \abbrDMS pops a label $l$ from OPEN.
\abbrDMS calls \procName{TargetSeq} for $l$, which takes the parent label of $l$ (denoted as $parent(l)$), $l$ itself, and the conflict set of $parent(l)$.
\procName{TargetSeq} either calls \procName{SolveMHPP} to find a joint sequence $\gamma(l)$, or re-uses the joint sequence $\gamma(parent(l))$ that is previously computed for $parent(l)$.
We elaborate \procName{TargetSeq} in Sec.~\ref{dms:sec:method:dts}.
After \procName{TargetSeq}, $\Vec{h}(l)$ may change since a joint sequence may be computed for $l$ within \procName{TargetSeq}, \abbrDMS thus computes $f_{max}(l)$ and compare it against $f_{temp}(l)$.
When computing $f_{max}$ out of $\vec{g}$ and $\vec{h}$ on Line~\ref{dms:alg:dms:lineFmax}, a heuristic inflation factor $w \in [1,\infty)$ is used, which scales each component in $\vec{h}$ by the factor $w$.
Heuristic inflation is common for \abbrAstar~\cite{pearl1984heuristics} and \abbrMstar-based algorithms~\cite{wagner2015subdimensional} that can often expedite the computation in practice while providing a $w$-bounded sub-optimal solution~\cite{pearl1984heuristics}.
If $f_{max}(l) > f_{temp}(l)$, then $l$ should not be expanded in the current iteration since there can be labels in OPEN that have smaller $f$-value than $f_{max}(l)$.
\abbrDMS thus updates $f_{temp}(l)$ to be $f_{max}(l)$, re-adds $l$ to OPEN with the updated $f_{temp}(l)$, and ends the iteration.
In a future iteration, when this label $l$ is popped again, the condition on Line~\ref{dms:alg:dms:lineFmaxCheck} will not hold since $f_{temp}(l)=f_{max}(l)$, and $l$ will be expanded.

Afterwards, \abbrDMS checks if $l$ leads to a solution using \procName{CheckSuccess}($l$), which verifies if every component in $\vec{a}(l)$ is one and if every component of $v(l)$ is a unique goal vertex while satisfying the assignment constraints.
If \procName{CheckSuccess}($l$) returns true, a solution joint path $\pi^*$ is found and can be reconstructed by iterative tracking the parent pointers of labels from $l$ to $l_o$ in \procName{Reconstruct}($l$).
\abbrDMS then terminates.
If \procName{CheckSuccess}($l$) returns false, $l$ is expanded by considering its limited neighbors~\cite{wagner2015subdimensional} described as follows.
The limited neighbors of $l$ is a set of successor labels of $l$.
For each $i \in I$, if $i \notin I_C(l)$, agent $i$ is only allowed to move to its next vertex $\phi_{\gamma(l)}^i(v^i(l))$ as defined in the joint policy $\phi_{\gamma(l)}$.
If $i \in I_C(l)$, agent $i$ is allowed to visit any adjacent vertex of $v^i(l)$ in $G^W$.
The successor vertices of $v^i(l)$ are:
\begin{gather}\label{dms:eqn:limited_succ}
u^i \gets
\begin{cases}
\phi^i_{\gamma(l)}(v^i(l)) &\mbox{if } i \notin I_C(l) \\
u^i \;|\; (v^i(l),u^i) \in E^W & \mbox{if } i \in I_C(l)
\end{cases}
\end{gather}
Let $V_{succ}^i$ denote the set of successor vertices of $v^i(l)$, which is either of size one or equal to the number of edges incident on $v^i(l)$ in $G^W$.
The successor joint vertices $V_{succ}$ of $v(l)$ is then the combination of $v^i(l)$ for all $i\in I$, i.e., $V_{succ}:= V_{succ}^1 \times V_{succ}^2 \times \cdots \times V_{succ}^N$.
For each joint vertex $v' \in V_{succ}$, a corresponding $l'$ is created and added to $L_{succ}$, the set of successor labels of $l$.
When creating $l'$, the corresponding $\vec{g}(l')$ and $\vec{a}(l')$ are computed based on $\vec{g}(l)$ $\vec{a}(l)$ and $v'$.
{\blue In other words, the element in $\vec{g}(l')$ is one unit larger than the corresponding element in $\vec{g}(l)$ (unless the agent has reached the goal and stays there) since every agents takes an action, and the element in $\vec{a}(l')$ changes its value from 0 to 1, if $v'$ visits any targets that are unvisited as in $a(l)$.}

After generating the successor labels $L_{succ}$ of $l$, for each $l' \in L_{succ}$, \abbrDMS checks for conflicts between agents during the transition from $v(l)$ to $v(l')$, and store the subset of agents in conflict in the conflict set $I_C(l')$.
\abbrDMS then invokes \procName{BackProp} (Alg.~\ref{dms:alg:backprop}) to back propagate $I_C(l')$ to its ancestor labels recursively so that the conflict set of these ancestor labels are modified, and labels with modified conflict set are re-added to OPEN and will be re-expanded.
\abbrDMS maintains a back\_set($l$) for each $l$, which is a set of pointers pointing to the predecessor labels to which the back propagation should be conducted.
Intuitively, similarly to~\cite{wagner2015subdimensional,ren21ms}, the conflict set of labels are dynamically enlarged during planning when agents are detected in conflict.
The conflict sets of labels determine the sub-graph within the joint graph $\mathcal{G^W}$ that can be reached by \abbrDMS, and \abbrDMS always attempts to limit the search within a sub-graph of $\mathcal{G^W}$ as small as possible.

Afterwards, if $I_C(l') \neq \emptyset$, $l'$ leads to a conflict and is discarded.
Otherwise, $l'$ is checked for pruning by using dominance (Def.~\ref{dms:def:labelDom}) against any existing labels in $\FrontierSet(v(l'))$.
If $l'$ is pruned, any future joint path from $l'$ can be cut and paste to $l'' \in \FrontierSet(v(l'))$ that dominates $l'$ without worsening the cost to reach the goals.
Furthermore, for each $l'' \in \FrontierSet(v(l'))$ that dominates or is equal to $l'$, \procName{DomBackProp}(Alg.~\ref{dms:alg:dom_back_prop}) is invoked so that the conflict set $I_C(l')$ is back propagated to $l$, and $l$ is added to the back\_set of $l''$.
By doing so, \abbrDMS is able to keep updating the conflict set of the predecessor labels of $l'$ after $l'$ is pruned.
This ensures the predecessor labels of $l'$ will also be re-expanded after $l'$ is pruned.
If $l'$ is not pruned, \procName{SimpleHeu} is invoked for $l'$ to compute a heuristic, which can be implemented by first copying $\vec{h}(l)$, where $l$ is the parent of $l'$, and then reduce each component of the copied vector by one except for the components that are already zero.
This heuristic is an underestimate of the cost-to-go since all agents can move at most one step closer to their goals in each expansion.
Then, $f_{temp}(l')$ is computed and $l'$ is added to OPEN with $f_{temp}(l')$ as its priority.
Other related data structure including $\FrontierSet(v(l'))$, back\_set, $parent$ are also updated correspondingly, and the iteration ends.

When \abbrDMS terminates, it either finds a conflict-free joint path, or returns failure when OPEN is empty if the given instance is unsolvable.

\subsection{Deferred Target Sequencing}\label{dms:sec:method:dts}
\abbrDMS introduces two techniques to defer the target sequencing until needed.
As aforementioned, the first one uses a fast-to-compute yet roughly estimated cost-to-go as the heuristic when a label is generated, and invokes \procName{TargetSeq} only when that label is popped from OPEN for expansion.

We now focus on the second technique in \abbrDMS.
In the previous \abbrMS~\cite{ren21ms}, every time when the search encounters a new label $l$ that is off-policy, inside \procName{TargetSeq}, \procName{SolveMHPP} is invoked for $l$ to find a joint sequence and policy from $l$, which burdens the overall computation, especially when a lot of new labels are generated due to the agent-agent conflict.
Different from \abbrMS, \abbrDMS seeks to defer the call of \procName{SolveMHPP} inside \procName{TargetSeq}.
For a label $l'$, \abbrDMS attempts to avoid calling \procName{SolveMHPP} for $l'$ by re-using the joint sequence $\gamma(l)$ of its parent label $l$ ($l$ is the parent of $l'$).
As presented in Alg.~\ref{dms:alg:targetSeq}, on Lines~\ref{dms:alg:targetSeq:lineCopySeq}-\ref{dms:alg:targetSeq:lineCopySeqThen}, \abbrDMS first attemps to build a policy from $l'$ by following the joint sequence of its parent label $\gamma(l)$ and computes the corresponding cost-to-go $\Vec{h}(l')$.
Agents $i \notin I_C(l)$ are still along their individual paths as in the previously computed policy $\phi_{\gamma(l)}$.
Agents $i\in I_C(l)$ consider all actions as in Equation (\ref{dms:eqn:limited_succ}) and may deviate from the individual paths specified by the previously computed policy $\phi_{\gamma(l)}$.
Therefore, \abbrDMS needs to go through a check for these agents $i\in I_C(l)$:
\abbrDMS first computes the $f$-vector by summing up $\vec{g}(l')$ and $\vec{h}(l')$.
Then, if $f^i(l')$ of some agent $i\in I_C(l)$ is no larger than $f_{max}(l)$, \abbrDMS can avoid calling \procName{SolveMHPP}, since letting the agents follow $\gamma(l)$ in the future will not worsen the objective value, the makespan.
Otherwise, there exists an agent $i\in I_C(l)$, whose corresponding $f^i(l')$ is greater than $f_{max}(l)$, and in this case, \abbrDMS cannot avoid calling \procName{SolveMHPP} since there may exists another joint sequence from $l'$, which leads to a solution joint path with better (smaller) objective value.

        \section{Analysis}\label{dms:sec:analysis}
        
\subsection{Completeness}
This section discusses the properties of \abbrDMS.
An algorithm is \emph{complete} for a problem if it finds a solution for any solvable instance, and reports failure in finite time for unsolvable instances.

\begin{theorem}
    If \procName{SolveMHPP} is complete, then \abbrDMS is complete for \abbrMCPFMax.
\end{theorem}

\begin{proof}
Let $\mathcal{A} = \{0,1\}^M$ denote the set of all possible binary vectors and let $\mathcal{S}=\mathcal{G^W}\times \mathcal{A}$ denote the Cartesian of the joint graph $\mathcal{G^W}$ and $\mathcal{A}$.
$\mathcal{S}$ is a finite space.
\abbrDMS uses dominance pruning and does not expand the same label twice during the search (the same label will be pruned by dominance).
The policies computed by \abbrDMS defines a sub-graph $\mathcal{G}^{W}_{sub}$ of the joint graph $\mathcal{G^W}$.
\abbrDMS first expands labels in this sub-graph $\mathcal{G}^{W}_{sub}$.
If no conflict is detected when following the policy, then the resulting joint path is conflict-free and solves the problem.
If there is a conflict, \abbrDMS may re-expand a label with a larger conflict set. The conflict set of a label can be enlarged for at most $(N-1)$ times.
Therefore, there is only a finite number of re-expansion of labels.
As a result, for a unsolvable instance, \abbrDMS enumerates all possible joint paths that starts from $(v_o,\vec{0}^M)$ to any reachable $(v,a) \in \mathcal{S}$ and terminates in finite time.
For a solvable instance, \abbrDMS conducts systematic search in $\mathcal{S}$ and returns a solution in finite time.
\end{proof}

The above analysis requires that the procedure \procName{SolveMHPP} is complete for MHPP. Otherwise, if \procName{SolveMHPP} never terminates for certain MHPP instances, \abbrDMS will not terminate in finite time when calling \procName{SolveMHPP} with those MHPP instances.

\subsection{Solution Optimality}
To discuss the solution optimality of \abbrDMS, we needs the following assumptions.

\begin{assumption}
    The procedure \procName{SolveMHPP} returns an optimal joint sequence for the given \abbrMHPP instance.
\end{assumption}

\begin{assumption}
    Line~\ref{dms:alg:targetSeq:lineCheckFmax} in Alg.~\ref{dms:alg:dms} never returned true during the search of \abbrDMS.
\end{assumption}

\begin{theorem}
    For a solvable instance, when Assumption 1 and 2 hold, \abbrDMS returns an optimal solution joint path.
\end{theorem}

The proof of Theorem 2 follows the analysis in~\cite{wagner2015subdimensional,ren21ms}, and we highlight the main ideas here.
The policies computed by \abbrDMS defines a sub-graph $\mathcal{G}^{W}_{sub}$ of the joint graph $\mathcal{G^W}$.
\abbrDMS first expands labels in this sub-graph $\mathcal{G}^{W}_{sub}$.
If no conflict is detected when following the policy, then the resulting joint path is conflict-free and optimal due to Assumption 1:
With Assumption 1, ${h}_{max}(l)$ for any label $l$ computed by \procName{SolveMHPP} is an estimated cost-to-go that is admissible, i.e., ${h}_{max}(l)$ is no larger than the true optimal cost-to-go.
This is true because \procName{SolveMHPP} ignores agent-agent conflicts and the resulting ${h}_{max}(l)$ must be a lower bound on the true cost-to-go.
As \abbrDMS selects labels from OPEN in the same way as \abbrAstar does, admissible heuristics lead to an optimal solution~\cite{pearl1984heuristics}.

If conflicts are detected, \abbrDMS updates (i.e., enlarges) $\mathcal{G}^{W}_{sub}$ by enlarging the conflict set and back propagating the conflict set.
When Assumption 2 holds, the enlarged $\mathcal{G}^{W}_{sub}$ still ensures that an optimal conflict-free joint path $\pi_*$ is contained in the updated $\mathcal{G}^{W}_{sub}$~\cite{wagner2015subdimensional,ren21ms}.
\abbrDMS systematically search over $\mathcal{G}^{W}_{sub}$ and finds $\pi_*$ at termination.

We now explain if Assumption 2 is violated, why \abbrDMS loses the optimality guarantee and how to fix it.
In \abbrMCPFMax, all agents are ``coupled'' in the space of binary vectors $\mathcal{A}$ in a sense that a target visited by one agent does not need to be visited by any other agents.
As as result, when one agent changes its target sequence by visiting some target that is previously assigned to another agent, all agents may need to be re-planned in order to ensure solution optimality.
It means that, when Line~\ref{dms:alg:targetSeq:lineCheckFmax} in Alg.~\ref{dms:alg:dms} returns true for label $l'$, procedure \procName{SolveMHPP} needs to be called for $l'$ and \procName{SolveMHPP} may return a new joint sequence $\gamma'$.
In this new sequence $\gamma'$, it is possible that agents within $I_C(l)$ visits targets that are previously assigned to agents that are not in $I_C(l)$.
However, \abbrDMS does not let agents outside $I_C(l)$ to take all possible actions as defined in Equation~(\ref{dms:eqn:limited_succ}).
As a result, an optimal solution joint path may lie outside $\mathcal{G}^{W}_{sub}$ and \abbrDMS thus loses the solution optimality guarantee when Assumption 2 is violated.

To ensure solution optimality when Assumption 2 is violated, same as in~\cite{ren21ms}, one possible way is to back propagate the entire index set $I=\{1,2,\cdots,N\}$ as the conflict set when calling \procName{BackProp}.
This ensures that \abbrDMS considers all possible actions for all agents when conflicts between agents are detected.
However, in practice, this is computationally burdensome when $N$ is large, and limits the scalability of the approach.
We therefore omit it from Alg.~\ref{dms:alg:dms} for clarity.

\subsection{Solution Bounded Sub-Optimality}

A joint sequence $\gamma$ is $\epsilon$-bounded sub-optimal ($\epsilon \geq 0$) if $\max_{i\in I}cost(\gamma^i) \leq (1+\epsilon) \max_{i\in I}cost(\gamma_*^i)$, where $\gamma_*$ is an optimal joint sequence.
Similarly, a solution joint path $\pi$ is $\epsilon$-bounded sub-optimal if its objective value $g \leq (1+\epsilon) g_*$, where $g_*$ is the true optimum of the given instance.

\begin{theorem}
    For a solvable instance, if (i) \procName{SolveMHPP} returns a $\epsilon$-bounded sub-optimal joint sequence for any given label $l$ and (ii) Assumption 2 holds, then \abbrDMS returns a $\epsilon$-bounded solution joint path.
\end{theorem}

\begin{proof}    
Bounded sub-optimal joint sequences lead to inflated heuristic values for labels in \abbrDMS.
As \abbrDMS select labels based on $f$-values as \abbrAstar does, inflated heuristic values lead to bounded sub-optimal
solutions for the problem~\cite{pearl1984heuristics}.
\end{proof}

In addition to using bounded sub-optimal joint sequences, \abbrDMS is also able to use the conventional heuristic inflation technique~\cite{pearl1984heuristics} and provide bounded sub-optimal solution.

\begin{theorem}
    For a solvable instance, if Line~\ref{dms:alg:dms:lineGetF} in Alg.~\ref{dms:alg:dms} is replaced with $\vec{f} \gets \vec{g} + (1+w)\vec{h}$ and Assumption 1 and 2 hold, then \abbrDMS returns a $w$-bounded solution joint path.
\end{theorem}

	\section{Experimental Results}\label{dms:sec:result}
	
We implement \abbrDMS in Python, with Google OR-Tool
as the \abbrMHPP solver.
Limited by our knowledge on Google OR-Tool, we only consider the following type of assignment constraints, where any agent can visit all targets and goals.
We set up a 60-second runtime limit for each instance, where each instance contains the starts, targets and goals in a grid map.
All tests run on a MacBook Pro with a Apple M2 Pro CPU and 16GM RAM.
It is known from~\cite{wagner2015subdimensional}, \abbrMstar-based algorithms with inflated heuristics can often scale to more agents, and we set $w=1.1$ in our tests.
We set the number of targets $M=20,40,60,80$.
We compare \abbrDMS, which includes the two proposed technique to defer target sequencing, against two baselines.
The first one is \abbrMS, which does not have these two techniques. In other words, \abbrMS here is a naive adaption of the existing \abbrMS~\cite{ren21ms} algorithm to solve \abbrMCPFMax, by using the aforementioned Google OR-Tool to solve min-max \abbrMHPP for target sequencing.
The second baseline is a iterative greedy approach which assigns one unvisited target to an agent to minimize the makespan of all agents in the current iteration while planning collision free paths.

\begin{figure}[tb]
\centering
\includegraphics[width=0.9\linewidth]{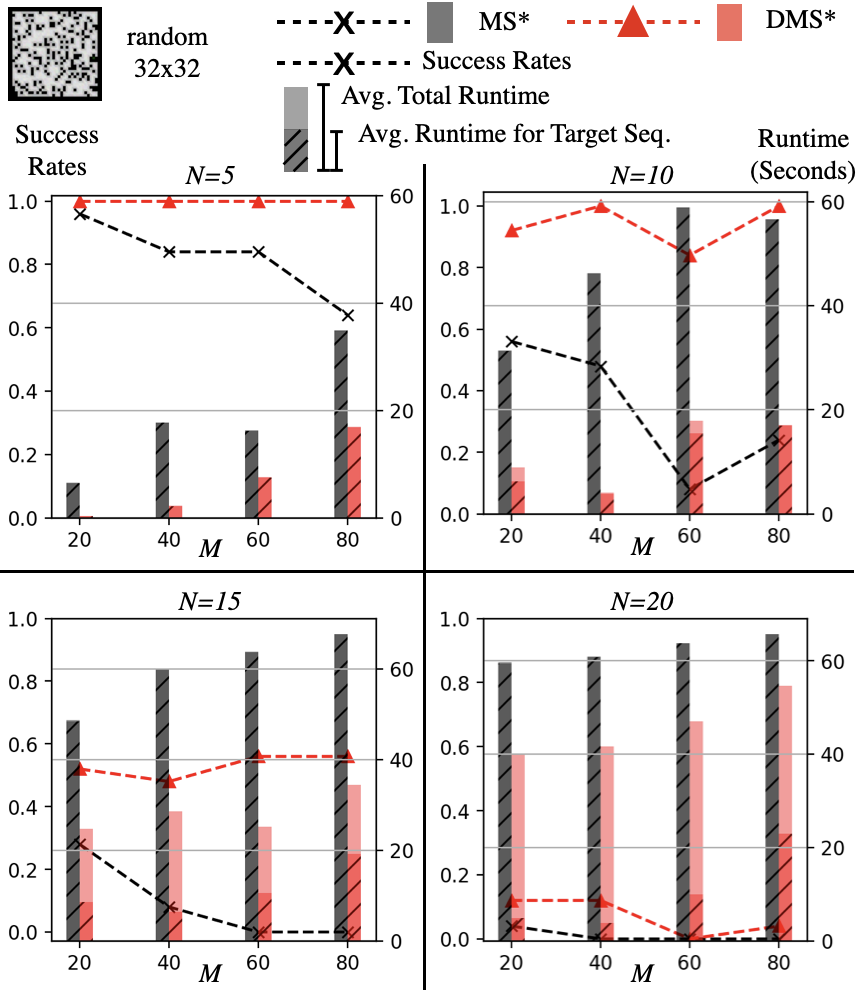}
\caption{
    The success rates and runtime of \abbrDMS and \abbrMS (baseline) with varying number of agents and targets in a random 32x32 map. \abbrDMS has higher success rates and less runtime on average than \abbrMS. 
    }
\label{dms:fig:resN}
\end{figure}

\begin{figure}[tb]
\centering
\includegraphics[width=0.9\linewidth]{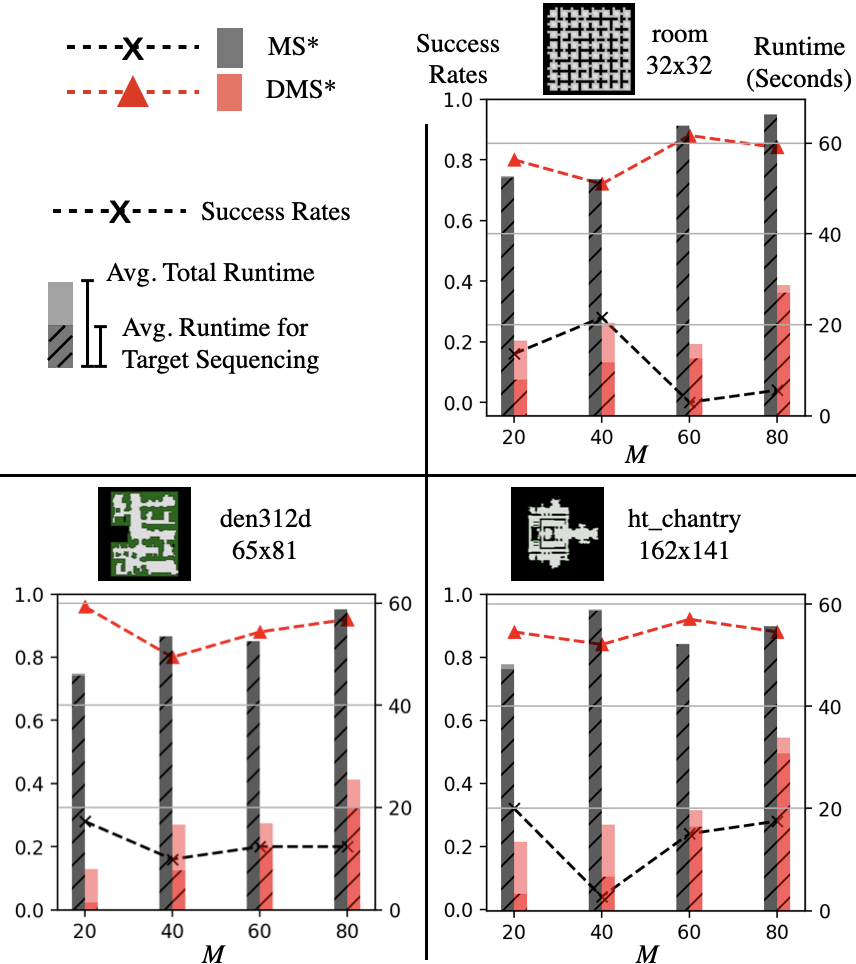}
\caption{
    The success rates and runtime of both \abbrDMS and \abbrMS (baseline) with varying number of targets in maps of different sizes. \abbrDMS achieves higher success rates than \abbrMS while requiring less runtime on average.
    }
\label{dms:fig:resMaps}
\end{figure}

\subsection{Simulation Results}

\subsubsection{Varying Number of Agents}

We first fix the map to Random 32x32, and vary the number of agents $N=5,10,15,20$.
We measure the success rates, the average runtime to solution and the average runtime for target sequencing per instance.
The averages are taken over all instances, including both succeeded instances and instances where the algorithm times out.
As shown in Fig.~\ref{dms:fig:resN}, \abbrDMS achieves higher success rates and lower runtime than the baseline \abbrMS.
As $M$ increases from $20$ to $80$, both algorithms require more runtime for target sequencing.
As $N$ increases from $5$ to $20$, agents have higher density and are more likely to run into conflict with each other.
As a result, both algorithms time out for more instances.
Additionally, \abbrMS spends almost all of its runtime in target sequencing, which indicates the computational burden caused by the frequent call of \abbrMHPP solver.
In contrast, for \abbrDMS, when $N=5,10$, \abbrDMS spends most of the runtime in target sequencing, while as $N$ increases to $15,20$, \abbrDMS spends more runtime in path planning.

\subsubsection{Different Maps}

We then fixed $N=10$ and test in maps of different sizes (Fig.~\ref{dms:fig:resMaps}).
\abbrDMS outperforms \abbrMS in success rates due to the alleviated computational burden for target sequencing.
Larger maps do not lead to lower success rates since larger maps can reduce the density of the agents and make the agents less likely to run into conflicts with each other.
The Room 32x32 map is more challenging than the other two maps since there are many narrow corridors which often lead to conflicts between the agents.

\begin{table}[tb]
    \centering
    \small
        \tabcolsep=0.1cm
    \begin{tabular}{ | l | l | l | l | }
        \hline
        $M$ & min. CR & median CR & max.CR
        \\ \hline
        $20$ & 0.47 & 0.71 & 0.94
        \\ \hline
        $40$ & 0.46 & 0.63 & 0.86
        \\ \hline
        $60$ & 0.53 & 0.66 & 0.76
        \\ \hline
        $80$ & 0.56 & 0.72 & 0.87
        \\ \hline
    \end{tabular}
    \caption{Cost ratios (CR) of \abbrDMS over a greedy baseline. Solution costs of \abbrDMS are up to 50\% cheaper than the greedy.}
    \label{dms:tab:cr}
\end{table}

\subsubsection{Solution Quality}

We use the random 32x32 map and fixed the number of agents to $N=5$ where \abbrDMS achieves 100\% success rate.
We compare the solution cost ratio of \abbrDMS over the greedy baseline.
As shown in Table~\ref{dms:tab:cr}, the solution of \abbrDMS is up to around 50\% cheaper than the solution of this greedy baseline.

\begin{table}[tb]
    \centering
    \small
        \tabcolsep=0.1cm
    \begin{tabular}{ | l | l | l | l | l | }
        \hline
        $M$ & 20 & 40 & 60 & 80
        \\ \hline
        \abbrMS & 72.2 & 67.6 & 75.0 & 71.8
        \\ \hline
        \abbrDMS & 77.2 & 84.4 & 93.4 & 99.9
        \\ \hline
    \end{tabular}
    \caption{The average number of expansion of \abbrMS and \abbrDMS among the instances where both algorithms succeed.}
    \label{dms:tab:exp}
\end{table}
{\blue
\subsubsection{Number of Expansions}
Table~\ref{dms:tab:exp} shows the average number of label expansion in both \abbrMS and \abbrDMS among the instances in the Random 32x32 map which are successfully solved by both algorithms within the runtime limit.
Due to the deferred sequencing, \abbrDMS tends to search in a less informed manner and needs more expansion.

}

\subsection{Experiments with Mobile Robots}

\ifthenelse{\boolean{shortver}}{%

We test our method with two differential drive robots, which are shown in the multi-media attachment.
This test verifies that the paths planned by DMS* are executable on real robots.
When the robots have large motion disturbance, such as delay or deviation from the planned path, additional techniques (such as~\cite{honig2016multi}) will be needed to ensure collision-free execution of the paths.

}{%
	
\begin{figure}[tb]
\centering
\includegraphics[width=\linewidth]{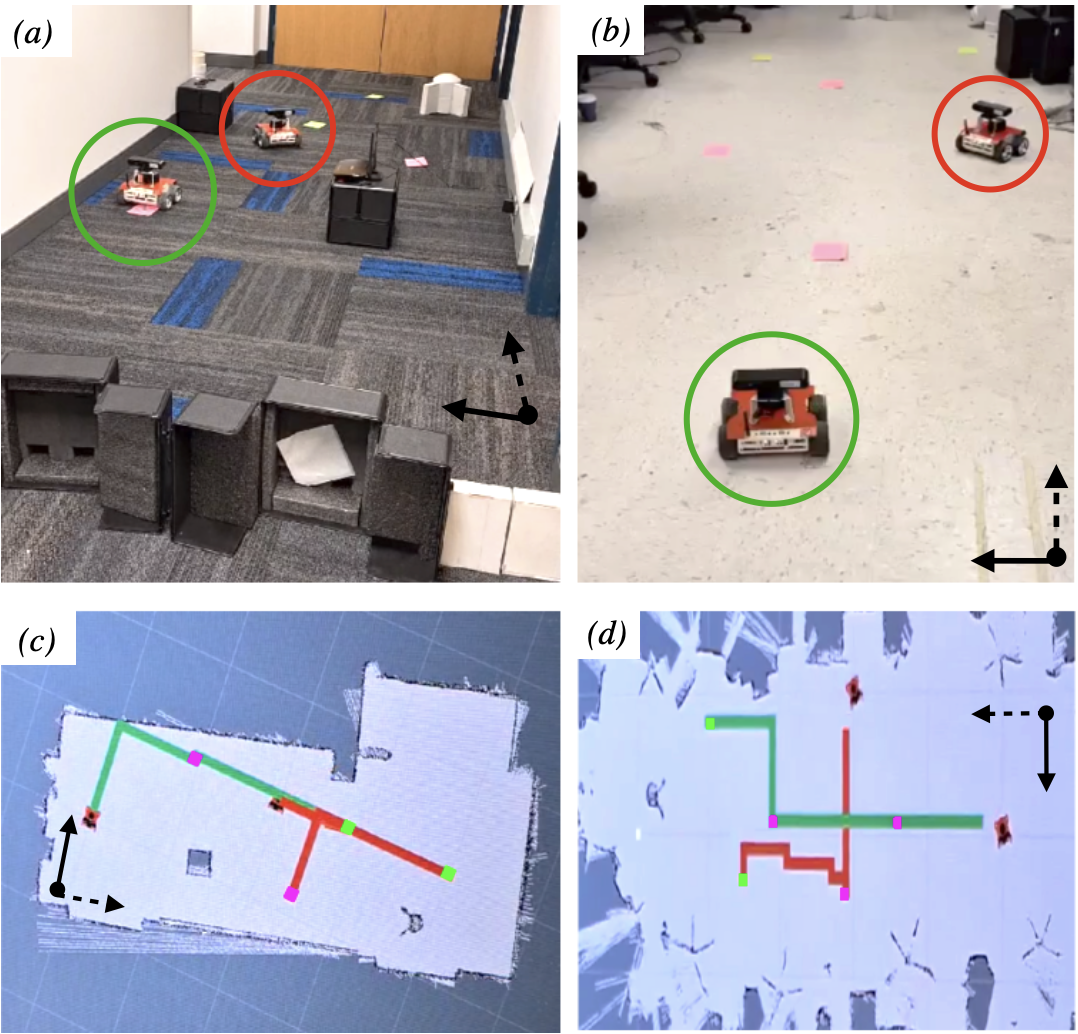}
\caption{Test settings for ROSbot experiments. (a) and (b) show the setup. (c) and (d) show the rviz window corresponding to (a) and (b) respectively. 
Targets are highlighted using pink boxes and the goals using green.}
\label{dms:fig:rosbot}
\end{figure}

\begin{figure}[htb]
\centering
\includegraphics[width=\linewidth]{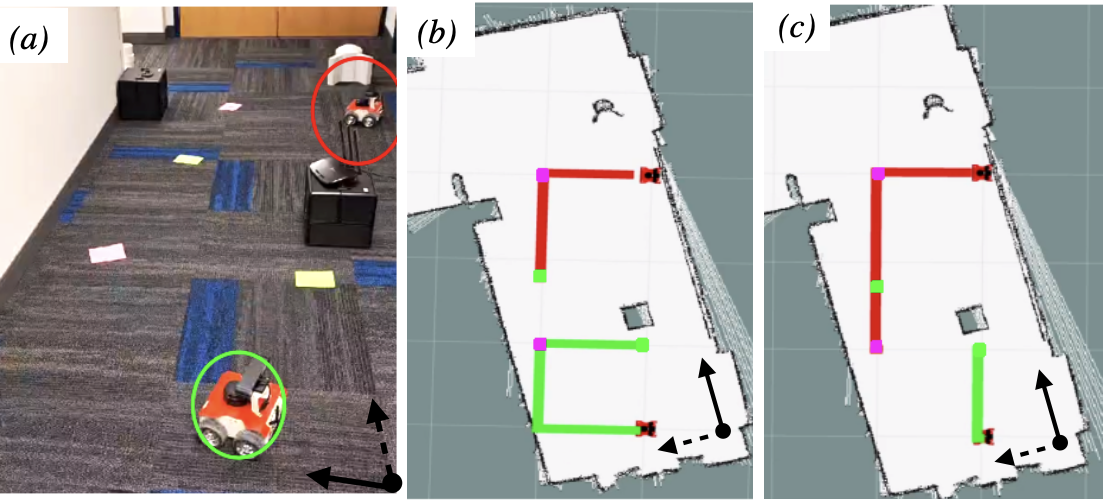}
\caption{Comparison of \abbrDMS and \abbrMS on mobile robots. (a) shows the test setting.
(b) and (c) show the joint path planned by \abbrDMS and \abbrMS respectively.
Both planners return different paths since they optimize different objective functions.}
\label{dms:fig:comparision_exp}
\end{figure}

We also performed experiments with {Husarion ROSbot 2}, which are differential drive robots equipped with an ASUS Tinker Board and use a 2D lidar for localization.
We used two such robots to perform tests in two different environments, as shown in Fig.~\ref{dms:fig:rosbot} (a) and (b) having 2 and 3 targets, respectively.
This test verifies that the paths planned by DMS* are executable on robots.
When the robots have large motion disturbance, such as delay or deviation from the planned path, additional techniques (such as~\cite{honig2016multi}) will be needed to ensure collision-free execution of the paths, which is another research topic and is not the focus of this work.
In our experiments, the robots run in slow speed and the motion disturbance of the robots are relatively small.

We then compare the performance of \abbrDMS and \abbrMS on the robots.
Both planners were tested on the same map with two intermediate targets as shown in Fig.~\ref{dms:fig:comparision_exp}.
We observe that \abbrDMS tends to assign the targets evenly to the agents and plan paths to minimize the makespan, while \abbrMS finds a solution wherein one robot visits all targets, and the other robot directly goes to its goal.
\abbrMS seeks to minimize the sum of arrival times, which can lead to a large makespan.

}
	
	\section{Conclusion and Future Work}\label{dms:sec:conclude}
	This paper investigates a min-max variant of Multi-Agent Combinatorial Path Finding problem and develops \abbrDMS algorithm to solve this problem.
We test \abbrDMS with up to 20 agents and 80 targets and conduct simple robot experiments to showcase the usage of \abbrDMS.

For future work, one can investigate {\blue extending \abbrDMS to the case where edges have non-unitary traversal times~\cite{andreychuk2022multi,ren21loosely}, or simultaneously optimizing both the maximum and the sum of individual arrival times~\cite{ren23matcpf}.}
We note from our experiments the disturbance in robot motion may affect the execution of the planned path, and one can develop fast online replanning version of \abbrDMS to handle the disturbance.

\bibliographystyle{IEEEtran}
\bibliography{ref}

\end{document}